\documentclass[10pt,twocolumn,letterpaper]{article}

\usepackage{cvpr}              

\usepackage[accsupp]{axessibility}
\usepackage{bm}
\usepackage{color}
\usepackage{graphicx}
\usepackage{amsmath}
\usepackage{amssymb}
\usepackage{booktabs}
\usepackage{xcolor}
\usepackage{multirow}
\usepackage[normalem]{ulem}
\newcommand\hl{\bgroup\markoverwith
  {\textcolor{yellow}{\rule[-.5ex]{2pt}{2.5ex}}}\ULon}

\usepackage[pagebackref,breaklinks,colorlinks]{hyperref}

\usepackage[capitalize]{cleveref}
\crefname{section}{Sec.}{Secs.}
\Crefname{section}{Section}{Sections}
\Crefname{table}{Table}{Tables}
\crefname{table}{Tab.}{Tabs.}

\def\cvprPaperID{6985} 
\def\confName{CVPR}
\def\confYear{2022}

\def\W{\mathbf{W}}

\def\V{{\cal V}}
\def\S{{\mathcal S}}
\def\X{{\mathcal X}}
\def\n{{\mathbf n}}

\def\x{{\mathbf x}}

\def\d{{\mathbf d}}

\def\V{{\mathbf V}}

\def\c{{\mathbf c}}
\def\p{{\mathbf p}}

\begin{document}

\title{Self-Supervised Arbitrary-Scale Point Clouds Upsampling via Implicit Neural Representation}
\author{Wenbo Zhao$^{1,2}$, Xianming Liu$^{1,2}$\thanks{Corresponding author: csxm@hit.edu.cn}, Zhiwei Zhong$^{1,2}$, Junjun Jiang$^{1,2}$, Wei Gao$^{3}$, Ge Li$^{3}$, Xiangyang Ji$^{4}$\\
$^{1}$Harbin Institute of Technology, $^{2}$Peng Cheng Laboratory\\ $^{3}$Peking University Shenzhen Graduate School, $^{4}$Tsinghua University\\}
\maketitle
\begin{abstract}

Point clouds upsampling is a challenging issue to generate dense and uniform point clouds from the given sparse input. Most existing methods either take the end-to-end supervised learning based manner, where large amounts of pairs of sparse input and dense ground-truth are exploited as supervision information; or treat up-scaling of different scale factors as independent tasks, and have to build multiple networks to handle upsampling with varying factors. In this paper, we propose a novel approach that achieves self-supervised and magnification-flexible point clouds upsampling simultaneously. We formulate point clouds upsampling as the task of seeking nearest projection points on the implicit surface for seed points. To this end, we define two implicit neural functions to estimate projection direction and distance respectively, which can be trained by two pretext learning tasks. Experimental results demonstrate that our self-supervised learning based scheme achieves competitive or even better performance than supervised learning based state-of-the-art methods. {The source code is publicly available at https://github.com/xnowbzhao/sapcu.}

\end{abstract}

\section{Introduction}
\label{sec:intro}

Point clouds serve as a popular tool to represent 3D data due to their flexibility and compactness in describing objects/scenes with complex geometry and topology. They can be easily captured by modern scanning devices, and have been widely used in many applications, such as autonomous driving, robotics, etc. However, due to the inherent limitations of 3D sensing technology, raw point clouds acquired from 3D scanners are usually sparse, occluded and non-uniform. In many downstream applications, such as surface reconstruction and understanding, dense point clouds are desired for representing shapes with richer geometric details. Accordingly, people turn to develop a computational approach, referred to as point clouds upsampling, which has attracted extensive attention in both industry and academia \cite{PUNet,MPU,PUGAN,PUGCN,SSPU,CVPR21_Disentangled}.

Unlike conventional grid based images, point clouds are irregular and unordered, which make point clouds upsampling a more challenging task than its 2D image counterpart. The goal of point clouds upsampling is two-fold: 1) generating a dense point set from the sparse input to provide richer details of the object; 2) generating a uniform and complete point set to cover the underlying surface faithfully. 

In recent years, deep neural networks based point clouds upsampling approaches emerge and become popular, which adaptively learn structures from data and achieve superior performance than traditional methods, such as optimization-based ones~\cite{alexa2003computing,huang2009consolidation,huang2013edge}. For instance, Yu \textit{et al.} \cite{PUNet} propose to learn multi-level features per point and expand the point set via a multi-branch convolution unit implicitly in feature space, which is then split to a multitude of features for reconstruction of an upsampled point set.  Wang \textit{et al.} \cite{MPU} propose to progressively train a cascade of patch-based upsampling networks on different levels of detail. Li \textit{et al.} \cite{PUGAN} apply generative adversarial network into point clouds upsampling, which constructs an up-down-up expansion unit in the generator for upsampling point features with error feedback and self-correction, and formulate a self-attention unit to enhance the feature integration. To better represent locality and aggregate the point neighborhood information, Qian \textit{et al.} \cite{PUGCN} propose to use a Graph Convolutional Network to perform point clouds upsampling.

In summary, the outlined deep learning based methods take a general approach: first design an upsampling module to expand the number of points in the feature space, then formulate losses to enforce the output points to be as close as possible to the ground truth dense points. However, these methods suffer from the following two limitations:

    \noindent \textbf{End-to-End Training.} These methods are trained in an end-to-end supervised learning manner, which requires a large amount of pairs of input sparse and ground-truth dense point sets as the supervision information. The training data is constructed by sampling from synthetic models, whose distributions are inevitably biased from that of the real-scanned data. This would lead the trained models to have poor generalization ability in real-world applications. Thus, it is more desirable to develop self-supervised or unsupervised point cloud upsampling schemes.

    \noindent\textbf{Fixed Upsampling Factor.} Due to resource constraints, such as display resolution and transmission bandwidth, the required upsampling factor is usually various. These existing methods treat upscaling of different scale factors as independent tasks, which train a specific deep model for a pre-defined factor and have to build multiple networks to handle upsampling with varying factors. This manner is clumsy, which increases both model complexity and training time significantly. Thus, it is more desirable to develop unified point cloud upsampling schemes that can handle arbitrary scale factor.

Some methods developed very recently \cite{liu2020spunet,SSPU,Meta-PU,TIP21_Flexible} investigate the above limitations and attempt to address them:
\begin{itemize}
    \item Regarding self-supervised point cloud upsampling, Liu \textit{et al.} \cite{liu2020spunet} propose the coarse-to-fine framework, which downsamples the input sparse patches into sparser ones and then exploits them as pairs of supervision information to perform end-to-end training. \cite{SSPU} proposes an end-to-end self-supervised learning manner, in which the loss functions enforce the input sparse point cloud and the generated dense one to have similar 3D shapes and rendered images. However, these two methods are still limited to a fixed upsampling factor. 
    \item Regarding arbitrary-scale upsampling, inspired by the counterpart Meta-SR in image \cite{MetaSR}, Ye \textit{et al.} \cite{Meta-PU} propose Meta-PU for magnification-flexible point cloud upsampling, in which the meta-subnetwork is learned to adjust the weights of the upsampling blocks dynamically.  Qian \textit{et al.} \cite{TIP21_Flexible} design a neural network to adaptively learn unified and sorted interpolation weights as well as the high-order refinements, by analyzing the local geometry of the input point cloud. However, these two methods still follow the end-to-end supervised learning manner, which need to construct a large-scale training set including ground truth dense point sets with scales within a wide range.
\end{itemize}

In this paper, we propose a novel and powerful point clouds upsampling method via implicit neural representation, which can achieve self-supervised and magnification-flexible upsampling simultaneously. Specifically, to get rid of the requirement of ground truth dense point clouds, we do not directly learn the mapping between the input sparse and output dense point sets. Alternatively, inspired by the notion that an implicit surface can be represented by signed distance function (SDF) \cite{SDF,mescheder2019occupancy,DeepSDF}, we turn to seek the nearest projection points on the object surface for given seed points through two implicit neural functions, which are used to estimate projection direction and distance respectively. The two function can be trained by two constructed pretext self-supervised learning tasks. In the way, as long as the seed points are sampled densely and uniformly, we can produce high-resolution point clouds that are dense, uniform and complete. To guarantee the uniformity of seeds sampling, we exploit equally-paced 3D voxels to divide the space of point cloud. Experimental results demonstrate our self-supervised learning based scheme achieves competitive or even better performance that supervised learning based state-of-the-art methods. The main contributions of this work are highlighted as follows:
\begin{itemize}
\item To the best of our knowledge, we are the first in the literature to simultaneously consider self-supervised and arbitrary-scale point clouds upsampling. 
\item We formulate point clouds upsampling as the task of seeking nearest projection points on the implicit surface for seed points, which can be done by two implicit neural functions trained by pretext tasks. From the generated dense point clouds, we can achieve arbitrary-scale upsampling by farthest point sampling.
\item Although our method is self-supervised, it produces high-quality dense point clouds that are uniform and complete, and achieves competitive objective performance and even better visual performance compared with state-of-the-art supervised methods.
\end{itemize}

\section{Method}
\label{sec:method}

Define $\mathcal{X} = \{\p_i\}_{i=1}^n \in \mathbb R^{n\times3}$ as the input sparse point cloud. For a desirable scaling factor $r$, we target to obtain a corresponding dense point cloud $\mathcal{Y} = \{\p_i\}_{i=1}^N\in \mathbb R^{N\times3}$ including $N = \lfloor r\times n \rfloor$ points. $\S$ is defined as the underlying surface of the dense point cloud. The high-resolution point cloud $\mathcal{Y}$ is required to be dense and uniform, as well as be able to handle occlusion and noise, \textit{i.e.}, to be complete and clean.  

Unlike the existing methods that take the end-to-end training framework, we do not directly learn the mapping between the input sparse and output dense point sets,  but instead seek the nearest projection point on the object  surface for a given seed point in a self-supervised manner. By densely and uniformly sampling seed points in the space, we can obtain dense and approximately uniform projection points, which can describe the underlying surface faithfully.
The proposed self-supervised point clouds upsampling strategy includes the four steps:
\begin{itemize}
    \item {Seeds Sampling.} We represent the geometric space of the point cloud by 3D voxel grid, from which we choose the centres of voxels that are close to the implicit surface $\mathcal{S}$ as the seed points.
    \item {Surface Projection.} For seed points,  we project them to the implicit surface $\mathcal{S}$ to obtain the projected points, which construct the generated dense point cloud.
    \item {Outliers Removal.} We further remove the projected points that are generated by far seed points to achieve cleaner point cloud.
    \item {Arbitrary-Scale Point Cloud Generation.} To obtain the desired upsampling factor, we adjust the number of vertices of the generated dense clouds by farthest point sampling.
\end{itemize} 

In the following, we introduce each step in detail.

\subsection{Seeds Sampling}

To obtain uniformly sampled seed points, given a point cloud, we divide the 3D space into equally spaced voxels $\{\V_{(x,y,z)}\}$, where $\V_{(0,0,0)}$ represents the voxel locating in the origin of the 3D Cartesian coordinate system. We define the resolution of a 3D voxel volume as $l\times l \times l$. The centre of a voxel $\V_{(x,y,z)}$ is thus $\c_{(x,y,z)} = [x+0.5*l, y+0.5*l, z+0.5*l]$. The centres of voxels are equally distributed in the space, which serve as good candidates of seed points. However, we do not use them all, but choose the ones that are closed to the underlying surface of the point cloud.

A reasonable principle to choose centres is according to their distances to the surface $\S$. We choose a centre $\c_{(x,y,z)}$ as the seed if its distance to the surface within a preset range: $\mathbb{\text{Dist}}(\c_{(x,y,z)}, \S) \in [D_l, D_u]$. The difficulty lies in that we cannot directly compute the distance $\mathbb{\text{Dist}}(\c_{(x,y,z)}, \S)$, since the underlying surface $\S$ is unknown. We propose an alternative strategy to approximate $\mathbb{\text{Dist}}(\c_{(x,y,z)}, \S)$. Specifically, from the input sparse point sets $\mathcal{X}$, we choose $M$ points that are nearest to $\c_{(x,y,z)}$, denoted as $\{\p_{c,1},\p_{c,2},\cdots,\p_{c,m}, \cdots,\p_{c,M}\}$ that are ordered from near to far. From these points, we can form a set of triangles $\{T_m = (\p_{c,1},\p_{c,2},\p_{c,m})\}_{m=3}^M$. We then perform the following approximation:
\begin{equation}
    \mathbb{\text{Dist}}(\c_{(x,y,z)}, \S) \approx \min \mathbb{\text{Dist}}(\c_{(x,y,z)}, t), t \in \{T_m\}_{m=3}^M
\end{equation}
where $t$ represents a point contained in the constructed triangles.
Finally, we obtain the seed points set $C$. By setting appropriate $l$, we can generate dense and uniformly distributed seed points.

\begin{figure*}
\centering
\includegraphics[width=0.8\linewidth]{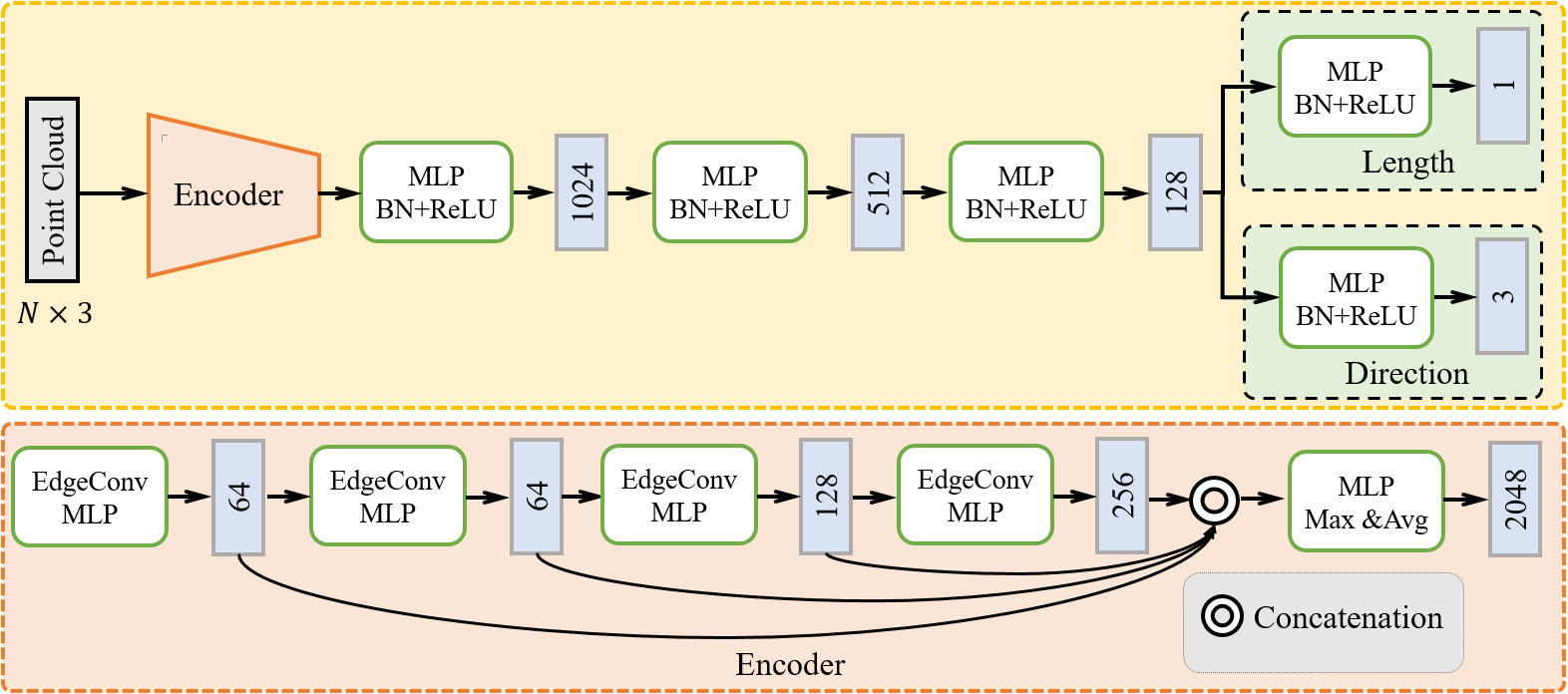}
    \caption{The network architecture of implicit neural function.
}
\label{c3}
\vspace{-0.3cm}
\end{figure*}

\subsection{Surface Projection}

With the sampled seed points, the next step is to seek the their projection points on the surface, which are the target points of the generated dense point cloud.

In the field of 3D computer vision and graphics, it is well-known that an implicit surface can be defined as a signed distance function (SDF) \cite{SDF,mescheder2019occupancy,DeepSDF}. SDF, when passed the coordinates of a point in space, outputs the closest distance of this point to the surface, whose sign indicates whether the point is inside or outside of the surface. Inspired by SDF, we propose the following feasible approach to estimate the projected point on the surface for a query seed point.

It is worth noting that, the computation strategies of SDF, such as \cite{SDF,mescheder2019occupancy,DeepSDF}, cannot be directly applied for our purpose. For a 3D query point $\x$, SDF outputs: $SDF(\x) = s, s\in \mathbb R$. The sign of $s$ only indicates it is inside or outside of a shape, but does not provide the direction to the surface.  In our method, for a seed point $\c \in C$, we divide the task of estimating projection point into two sub-tasks: 1) estimating the projection direction $\n\in[-1,1]^3$; 2) estimating the projection distance $d \in \mathbb R$. Then the coordinate of the projection point of the seed point $\c$ can be obtained as: $\c_p = \c + \n*d$.

\noindent \textbf{Projection Direction Estimation.} We train a multi-layer fully-connected neural network $f_n(\cdot;\Theta_n)$ for this purpose, which takes the query point $\c$ and the sparse point cloud $\mathcal X$ as inputs:
\begin{equation}
    \n = f_n (\c, \mathcal X;\Theta_n)
\end{equation}

To reduce the computational complexity, we take $k$ nearest points to $\c$ in $\mathcal X$  instead of the whole $\mathcal X$ as input. We denote this subset of points as $\X_c = \{\p_1,\cdots,\p_k\}$. Moreover, to facilitate the inference process of neural networks, we perform normalization on the point coordinates by setting $\c$ as the origin. In this way, we can simplify the estimation function as: 
\begin{equation}
    \n = f_n (\widehat{\mathcal X}_c;\Theta_n)
\end{equation}
where $\widehat \X_c = \{\p_1-\c,\cdots,\p_k-\c\}$.

\noindent \textbf{Projection Distance Estimation.} Similarly, for estimating the projection distance $\d$, we also train a multi-layer fully-connected neural network $f_d(\cdot;\Theta_n)$,  which takes the query point $\c$, the subset of nearest points ${\mathcal X}_c$ and the estimated projection direction $\n$ as inputs:
\begin{equation}
    d = f_d (\c, {\mathcal X}_c, \n;\Theta_d)
\end{equation}

Normalization is also helpful for this network. Different from $f_n$, here the input $\n$ involves direction. Therefore, it should perform normalization on both position and direction, which can be done in two stages:
1) moving $\c$ to the origin; 2) applying the rotation matrix $\W_r$ to rotate $\n$ to a specific direction $\n_t$, \textit{i.e.}, $\n_t = \W_r\n$. After normalization, ${\mathcal X}_c$ becomes $\widetilde{\mathcal X}_c =  \{\W_r(\p_1-\c),\cdots,\W_r(\p_k-\c)\}$, which is the only required input for $f_d$:
\begin{equation}
    d = f_d (\widetilde{\mathcal X}_c;\Theta_d)
\end{equation}

\subsection{Outliers Removal}

In the step of seeds sampling, some points that are actually far away from $\S$ may be included into the seed points set $C$ due to the error in approximation. The normal vector and distance of these points cannot be well estimated, leading to outliers in the resulting dense point cloud. We turn to exploit the post-processing procedure to remove them. 

Specifically, for a projection point $\c_p$, we find its $v$ nearest points $\{\c_{p,1}, \cdots, \c_{p,v}\}$. We then compute the average bias between $\c_p$ and them:
\begin{equation}
 b_p  = \frac{1}{v}\sum_{i=1}^v \text{Dist}(\c_p, \c_{p,i}) 
\end{equation}
For all projection points, we do in the same way to get $\{b_p\}$, the average of which is denoted as $\bar b$. We determine a point as outlier if it satisfies $b_p > \lambda \bar b$, where $\lambda$ is set as 1.5 in practical implementation.

\subsection{Arbitrary-Scale Point Cloud Generation}

Note that the above process cannot accurately control the number of vertices generated. Thus, it is necessary to adjust the number of vertices to achieve upsampling with the desired scale factor. In our context, we first perform inverse normalization on the generated point cloud, and then adjust the number of vertices to $N$ by the farthest point sampling algorithm \cite{pointnet++}.

\begin{table*}

\begin{center}
\renewcommand{\arraystretch}{1.2}
\footnotesize
\begin{tabular}{c|c||c|c|c|c|c||c|c|c|c|c}
\hline
  \multicolumn{2}{c||}{Scale} & \multicolumn{5}{c||}{2$\times$} & \multicolumn{5}{c}{4$\times$}  \\
\cline{1-12}
   \multicolumn{2}{c||}{Metric ($10^{-3}$)} & CD $\downarrow$	& EMD $\downarrow$ 	&  F-score $\uparrow$	& mean $\downarrow$	& std $\downarrow$ & CD $\downarrow$	&  EMD $\downarrow$	& F-score $\uparrow$ & mean $\downarrow$	& std $\downarrow$\\
\cline{1-12}
\multirow{5}*{\shortstack{Fixed\\Scale}}&PU-Net \cite{PUNet}      & 12.9 (6)& 6.75 (6)& 334 (5)& 4.02 (5)& 5.62 (4)&  11.3 (7)& 7.02 (7)& 462 (7)& 5.05 (7)& 6.81 (6)\\
&MPU \cite{MPU}       & - & -& -& -& -&                                   10.4 (4)& 5.64 (5)& 527 (4)& 3.61 (4)& 5.50 (3)\\
&PU-GAN \cite{PUGAN}    & 12.7 (5)& 5.09 (5)& 326 (6)& 4.32 (6)& 6.01 (5)&  10.9 (5)& 6.66 (6)& 484 (6)& 4.66 (6)& 6.56 (5)\\
&PU-GCN \cite{PUGCN}    & 12.2 (3)& 4.94 (4)& 360 (3)& 3.47 (2)& 5.09 (2)&  10.2 (3)& 5.50 (4)& 537 (3)& 3.35 (2)& 5.01 (2)\\
&PU-DR \cite{CVPR21_Disentangled}      & 11.0 (1)& 3.55 (1)& 409 (1)& 2.30 (1)& 4.05 (1)&  8.71 (1)& 3.98 (2)& 625 (1)& 2.24 (1)& 3.89 (1)\\
\cline{1-12}
\cline{1-12}
\multirow{2}*{\shortstack{Arbitrary\\Scale}}&Meta-PU  \cite{Meta-PU}   & 12.6 (4)& 3.94 (2)& 339 (4)& 3.77 (4)& 5.41 (3)&  10.9 (5)& 3.56 (1)& 506 (5)& 3.89 (5)& 5.60 (4)\\
&Proposed       & 12.1 (2)& 4.93 (3)& 371 (2)& 3.49 (3)& 10.2 (6)&  10.1 (2)& 4.87 (3)& 561 (2)& 3.49 (3)& 9.35 (7)\\
\cline{1-12}

\hline
  \multicolumn{2}{c||}{Scale} & \multicolumn{5}{c||}{$8\times$} & \multicolumn{5}{c}{$16\times$}  \\
\cline{1-12}
   \multicolumn{2}{c||}{Metric ($10^{-3}$)} & CD $\downarrow$	& EMD $\downarrow$ 	&  F-score $\uparrow$	& mean $\downarrow$	& std $\downarrow$ & CD $\downarrow$	&  EMD $\downarrow$	& F-score $\uparrow$ & mean $\downarrow$	& std $\downarrow$\\
\cline{1-12}
\multirow{5}*{\shortstack{Fixed\\Scale}}&PU-Net \cite{PUNet}      &9.67 (5)& 8.91 (6)& 611 (6)& 4.85 (6)& 6.81 (5)& 9.25 (7)& 10.4 (7)& 632 (7)& 6.04 (7)& 8.04 (6)\\
&MPU \cite{MPU}    & -       & -       & -      & -       & -       & 8.12 (5)& 7.74 (6)& 727 (5)& 4.01 (6)& 6.11 (5)\\
&PU-GAN \cite{PUGAN}     &8.82 (4)& 5.05 (3)& 676 (4)& 3.76 (4)& 5.51 (3)& 7.52 (2)& 6.02 (4)& 770 (3)& 3.14 (2)& 4.92 (2)\\
&PU-GCN \cite{PUGCN}  &8.78 (3)& 6.41 (5)& 678 (3)& 3.33 (2)& 5.06 (2)& 7.80 (4)& 7.44 (5)& 749 (4)& 3.39 (4)& 5.11 (3)\\
&PU-DR \cite{CVPR21_Disentangled}    &8.34 (1)& 3.95 (1)& 708 (1)& 2.99 (1)& 4.97 (1)& 7.29 (1)& 4.51 (1)& 779 (1)& 2.92 (1)& 4.72 (1)\\
\cline{1-12}
\multirow{2}*{\shortstack{Arbitrary\\Scale}}&Meta-PU  \cite{Meta-PU}      &9.71 (6)& 4.33 (2)& 625 (5)& 3.95 (5)& 5.68 (4)& 8.96 (6)& 5.62 (2)& 676 (6)& 3.87 (5)& 5.59 (4)\\
&Proposed      &8.70 (2)& 5.53 (4)& 706 (2)& 3.48 (3)& 8.84 (6)& 7.65 (3)& 5.98 (3)& 772 (2)& 3.35 (3)& 8.34 (7)\\
\cline{1-12}

\end{tabular}
\end{center}
\vspace{-0.3cm}
\caption{Objective performance comparison with respect to CD, EMD, F-score, mean and std with state-of-the-art methods. The ranking numbers are also provided.}
\label{table1}
\vspace{-0.5cm}
\end{table*}

\section{Implicit Neural Networks Training}

In this section, we introduce the architecture of implicit neural networks and the training strategy.

\subsection{Architectures}
The networks $f_n$ and $f_d$ share the same architecture as shown in Figure~\ref{c3}, which borrows the idea of encoder-decoder framework~\cite{mescheder2019occupancy}. The network takes the normalized subset of points as input, which are feed into the encoder to obtain a 2048-dimensional feature vector. Here we employ a state-of-the-art method DGCNN~\cite{wang2019dynamic} as the encoder to preserve surface information in multi-levels. The feature vector is then passed through 4 full-connected (FC) layers with batch normalization and ReLU, the output dimensional of which are 1024, 512 and 128 respectively. The output dimensional of the last FC layer is 3 for the projection direction $\n$ and 1 for the the projection distance $d$. 
Note that the design of the network is not the main contribution of this paper. We can exploit any suitable network for our purpose.

\subsection{Training Data Preparation}
To train the two implicit neural functions, we construct two pretext tasks, for which we prepare training samples that consist of 3D points and the corresponding ground truth projection direction and distance values. We train with normalized watertight meshes that are constructed by {the TSDF-Fusion presented in \cite{Stutz2018ARXIV, mescheder2019occupancy} from a subset of the ShapeNet \cite{chang2015shapenet} that consists of 13 major categories.}  Since $f_n$ and $f_d$ are designed with different purposes, we prepare different training pairs for them:

\begin{itemize}
  \item {For preparing the training data of $f_n$, we firstly generate the seed points: 50K seed points are randomly selected around the mesh surface by limiting that the distances between them and the surface are within a preset range $[D_l-\epsilon, D_u+\epsilon]$, where $\epsilon$ is introduced to increase robustness. For a seed point $\c$, we find the nearest point on the mesh and sample 5 points around it, then compute the average vector $\d$ between $\c$ and them. In this way, we can effectively handle the error in mesh reconstruction. The ground truth $\widehat\n$ is finally derived as the normalization of $\d$. Secondly, for every 16 seed points, we randomly select  2048 points as the corresponding sparse point cloud $\mathcal{X}$.}

    \item {For preparing the training data of $f_d$, 50K seed points are randomly selected around the mesh surface in the same way as $f_n$. We then find the corresponding nearest projection points on the surface. The distance between a seed point and the projection point is used as the ground truth $\widehat{d}$. The generation of $\mathcal{X}$ is in the same way as $f_n$.}
\end{itemize}     

{It is worth noting that, although the training data are 
generated from watertight meshes, our scheme is capable to handle non-watertight point clouds, which can be observed in the real-world cases shown in Figure~\ref{m5}.}

\subsection{Training Details}
The training of $f_n$ and $f_d$ is done by minimizing the sum over losses between the predicted and real direction/length values under the mean squared error (MSE) loss function.

The training process is conducted on a server with two Tesla V100 GPUs. The networks are trained for 1200 epochs with a batch size of 64, using the Adam algorithm. Following \cite{mescheder2019occupancy}, the learning rate is set as $10^{-4}$, and the other hyperparameters of Adam are set as $\beta_1=0.9$, $\beta_2=0.999$, $epsilon=10^{-8}$, weight decay $= 0$.

\section{Experiments}
\label{sec:experiments}
In this section, we provide extensive experimental results to demonstrate the superior performance of our method. 
\subsection{Comparison Study}
We compare the proposed self-supervised arbitrary-scale point clouds upsampling (SSAS) method with several state-of-the-art works, which can be divided into two categories in term of the scale factor: 1) fixed scale methods, including PU-Net~\cite{PUNet},  MPU~\cite{MPU},  PU-GAN~\cite{PUGAN},  PU-GCN~\cite{PUGCN},  PU-DR~\cite{CVPR21_Disentangled}; 2) arbitrary scale method, including Meta-PU~\cite{Meta-PU}.
Note that these methods are all supervised learning based. The compared models are trained with the released codes by their authors, following the default settings in their papers. 

We train all these compared methods following the approach mentioned in \cite{Meta-PU} for fair comparison. The test samples are from the dataset adopted by~\cite{PUNet,Meta-PU}. We {non-uniformly} sample 2048 points using Poisson disk sampling from 20 test models to form the test set.

\subsection{Parameters Setting}
The side length of voxel $l$ and the range of the distance between seed point and surface $[D_l, D_u]$ decide the number of seed points. However, the number is also depended on the shape of input point cloud. To ensure that enough points are generated, we set $l=0.004$ and $\left[D_l, D_u\right]=\left[0.011,0.015\right]$. Under this condition, the minimum number of generated points is 99001 (\textit{Chair}), which meets the demands of 16$\times$ or higher scale upsampling. The specific direction $\n_{t}$ in normalization can be arbitrarily chosen. We set $\n_{t}=\left(1,0,0\right)$ in our experiments. The number of nearest points $k$ is set as 100 in Section 2.2 and 30 in Section 2.3. The number of nearest point $M$ for computing $\mathbb{\text{Dist}}(\c_{(x,y,z)}, \S)$ affects the number of outliers and continuity of seed points. We set $M=10$ and further discuss the effect of different $M$ in ablation study.  

\begin{table*}
\begin{center}
\renewcommand{\arraystretch}{1.2}
\footnotesize
\begin{tabular}{c|c||c|c|c|c|c||c|c|c|c|c}
\hline
  \multicolumn{2}{c||}{Scale} & \multicolumn{5}{c||}{$2\times$} & \multicolumn{5}{c}{$4\times$}  \\
\cline{1-12}
 \multicolumn{2}{c||}{$p$ ($10^{-2}$)}   & 0.2\% $\downarrow$	& 0.4\% $\downarrow$	&  0.6\% $\downarrow$	& 0.8\%	$\downarrow$& 1.0\% $\downarrow$& 0.2\% $\downarrow$	& 0.4\% $\downarrow$	&  0.6\% $\downarrow$	& 0.8\%	$\downarrow$&
 1.0\% $\downarrow$\\
\cline{1-12}
\multirow{5}*{\shortstack{Fixed\\Scale}}&PU-Net \cite{PUNet}     &3.05 (6)& 2.33 (6)& 2.03 (6)& 1.86 (6)& 1.75 (6)& 2.72 (7)& 2.19 (7)& 1.97 (7)& 1.84 (7)& 1.76 (7)\\
&MPU \cite{MPU}                                             &   - & -& -& -& -&                                2.53 (6)& 2.03 (6)& 1.80 (6)& 1.67 (6)& 1.59 (6)\\
&PU-GAN \cite{PUGAN}                                          &2.46 (3)& 1.86 (3)& 1.62 (4)& 1.50 (4)& 1.43 (4)& 2.45 (4)& 1.94 (5)& 1.73 (5)& 1.62 (5)& 1.56 (5)\\
&PU-GCN \cite{PUGCN}                                          &2.68 (5)& 2.06 (5)& 1.80 (5)& 1.65 (5)& 1.57 (5)& 2.40 (3)& 1.93 (4)& 1.72 (4)& 1.61 (4)& 1.54 (4)\\
&PU-DR \cite{CVPR21_Disentangled}                                            &1.83 (1)& 1.34 (1)& 1.17 (1)& 1.09 (1)& 1.06 (1)& 1.77 (2)& 1.46 (2)& 1.34 (2)& 1.28 (2)& 1.24 (2)\\
\cline{1-12}
\multirow{2}*{\shortstack{Arbitrary\\Scale}}&Meta-PU  \cite{Meta-PU}      &2.53 (4)& 1.86 (3)& 1.58 (3)& 1.43 (3)& 1.35 (3)& 2.50 (5)& 1.87 (3)& 1.60 (3)& 1.45 (3)& 1.37 (3)\\
&Proposed       &1.96 (2)& 1.50 (2)& 1.33 (2)& 1.25 (2)& 1.21 (2)& 1.72 (1)& 1.40 (1)& 1.27 (1)& 1.21 (1)& 1.19 (1)\\
\cline{1-12}
  \multicolumn{2}{c||}{Scale} & \multicolumn{5}{c||}{$8\times$} & \multicolumn{5}{c}{$16\times$}  \\
\cline{1-12}
 \multicolumn{2}{c||}{p ($10^{-2}$)}   & 0.2\% $\downarrow$	& 0.4\% $\downarrow$	&  0.6\% $\downarrow$	& 0.8\%	$\downarrow$& 1.0\% $\downarrow$& 0.2\% $\downarrow$	& 0.4\% $\downarrow$	&  0.6\% $\downarrow$	& 0.8\%	$\downarrow$&
 1.0\% $\downarrow$\\
\cline{1-12}
\multirow{5}*{\shortstack{Fixed\\Scale}}&PU-Net \cite{PUNet}    &2.64 (5)& 2.21 (6)& 2.02 (6)& 1.91 (6)& 1.84 (6)& 2.86 (6)& 2.42 (6)& 2.22 (7)& 2.10 (7)& 2.02 (7)\\
&MPU \cite{MPU}                                             &-       & -       & -      & -       & -        & 2.30 (4)& 1.96 (4)& 1.82 (4)& 1.73 (4)& 1.69 (4)\\
&PU-GAN \cite{PUGAN}                                          &1.72 (2)& 1.47 (3)& 1.38 (3)& 1.34 (3)& 1.32 (3)& 1.79 (3)& 1.57 (3)& 1.48 (3)& 1.44 (3)& 1.41 (3)\\
&PU-GCN \cite{PUGCN}                                          &2.31 (4)& 1.93 (4)& 1.75 (4)& 1.65 (5)& 1.59 (5)& 2.42 (5)& 2.07 (5)& 1.91 (5)& 1.82 (5)& 1.76 (5)\\
&PU-DR \cite{CVPR21_Disentangled}                                            &1.42 (1)& 1.20 (1)& 1.13 (1)& 1.11 (1)& 1.11 (1)& 1.56 (1)& 1.38 (1)& 1.32 (1)& 1.29 (1)& 1.28 (1)\\
\cline{1-12}
\multirow{2}*{\shortstack{Arbitrary\\Scale}}&Meta-PU  \cite{Meta-PU}      &2.72 (6)& 2.05 (5)& 1.76 (5)& 1.60 (4)& 1.51 (4)& 3.27 (7)& 2.51 (7)& 2.14 (6)& 1.93 (6)& 1.79 (6)\\
&Proposed       &1.72 (2)& 1.45 (2)& 1.34 (2)& 1.29 (2)& 1.27 (2)& 1.75 (2)& 1.51 (2)& 1.41 (2)& 1.36 (2)& 1.33 (2)\\
\cline{1-12}
\hline

\end{tabular}
\end{center}

\caption{Uniformity performance comparison with respect to NUC scores. The ranking numbers are also provided.}
\label{table2}
\end{table*}

\begin{figure*}
\begin{center}
\includegraphics[width=0.98\linewidth]{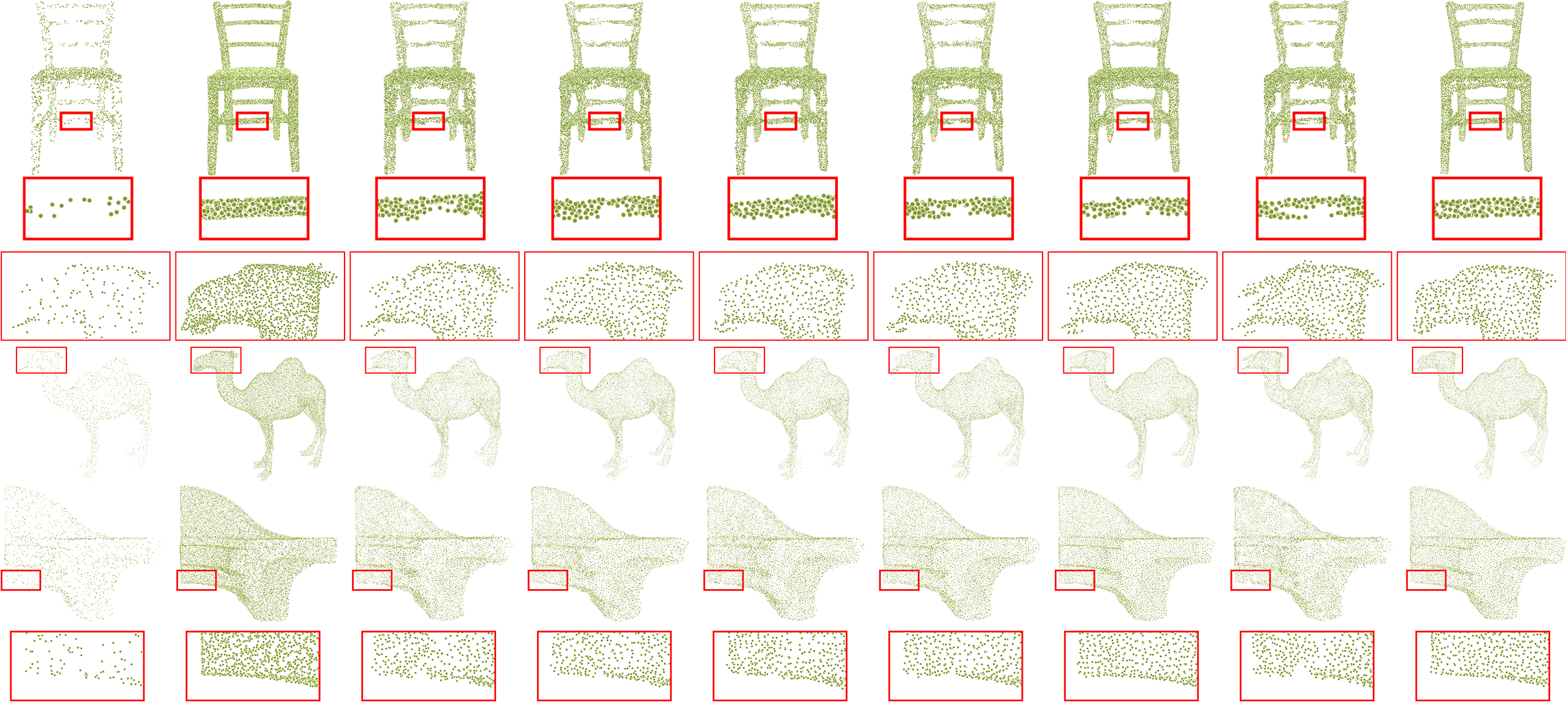}
{\hspace{0cm}(a)\hspace{1.54cm}(b)\hspace{1.54cm}(c)\hspace{1.54cm}(d)\hspace{1.54cm}(e)\hspace{1.54cm}(f)\hspace{1.54cm}(g)\hspace{1.54cm}(h)\hspace{1.54cm}(i)}
\end{center}
   \caption{$4\times$ point upsampling results of \textit{Chair}, \textit{Camel} and \textit{Fandisk}. (a) the input point cloud; (b) the ground truth;  (c) to (i) the results of PU-Net \cite{PUNet} , MPU \cite{MPU}, PU-GAN \cite{PUGAN}, PU-GCN \cite{PUGCN}, PU-DR \cite{CVPR21_Disentangled}, Meta-PU \cite{Meta-PU} and ours. Please enlarge the PDF for more details.
}
\label{m1}
\vspace{-0.3cm}
\end{figure*}

\subsection{Objective Performance Comparison}

 \noindent\textbf{Objective Evaluation.} We employ six popular metrics for objective evaluation:  1) \textbf{Chamfer Distance (CD)} and \textbf{Earth Mover Distance (EMD)}~\cite{PUNet}: which evaluate the similarity between the predicted points and ground truth one in the Euclidean space. For both metrics, smaller is better.
2) \textbf{F-score}~\cite{wu2019point}: which treats upsampling as a classification problem. For this metric, larger is better.
3) \textbf{mean} and \textbf{std}~\cite{PUNet}: which evaluate the distance between the predicted point cloud and ground truth mesh. For both metrics, smaller is better.
4) \textbf{Normalized Uniformity Coefficient (NUC)}~\cite{PUNet}: which evaluates the uniformity of points on randomly selected disk with different area percentage $p=0.2\%, 0.4\%, 0.6\%, 0.8\%, 1.0\%$. For this metric, smaller is better.

In Table \ref{table1}, we offer the comparison results for four scale factors $\left[2\times,4\times,8\times,16\times\right]$ with respect to CD, EMD, F-score, mean and std. Surprisingly, it can be found that, although our model is trained in a self-supervised manner without accessing to the ground-truth, it achieves competitive performance with those supervised learning based ones with respect to metrics CD, EMD, F-score and mean. Taking CD for example, our method is ranked \#2, \#2, \#2 and \#3 among seven compared methods for $\left[2\times,4\times,8\times,16\times\right]$ respectively. Similar results can be found for F-score, for which our method is ranked \#2 for all cases. Note that our method performs worst with respect to std, which is because that outliers cannot be completely removed.

Table \ref{table2} shows the uniformity evaluation results with respect to NUC. Our method is ranked \#2 for $\left[2\times,8\times,16\times\right]$, just blow PU-DR \cite{CVPR21_Disentangled}. In the case of $4\times$, our method is ranked \#1. These results demonstrate that the proposed method produces dense and uniform point clouds.

 \noindent\textbf{Inference Time Cost Comparison.} This experimental analysis is conducted on a server with two 1080Ti GPU. In our inference process, estimation of direction and length are the most time-consuming steps. According to the experiments, generating 40000 projected points by our method would cost 46s in average. As a comparison, the time costs of generating 40000 points ($16\times$) are 342.4s by MPU~\cite{MPU} and 0.228s by Meta-PU~\cite{Meta-PU}. It should be noticed that the estimation of each point is independently performed and thus can be done in parallel to speed up significantly.
\subsection{Subjective Performance Comparison}

\begin{figure}
\begin{center}
\includegraphics[width=0.98\linewidth]{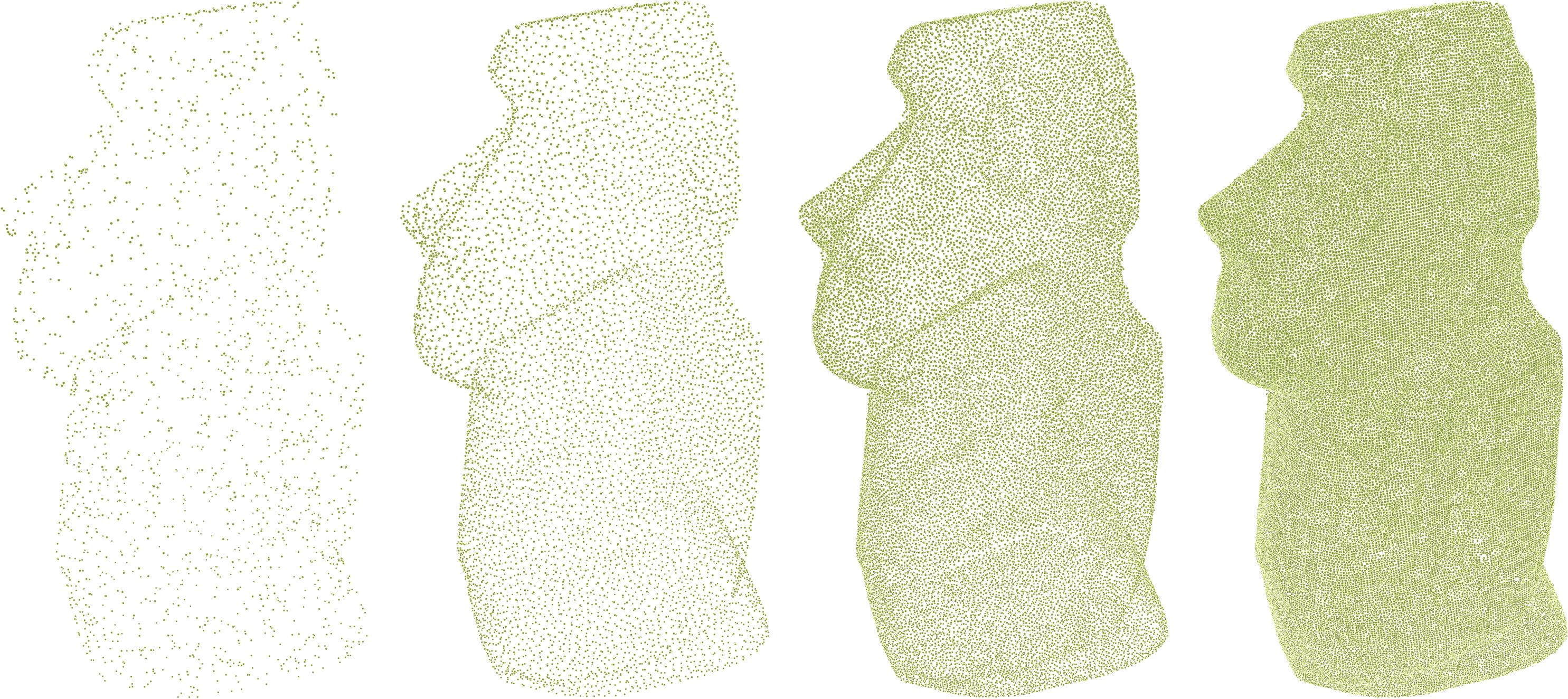}
{\hspace{0cm}Input\hspace{1.5cm}$4\times$ \hspace{1.5cm} $16\times$ \hspace{1.5cm} $64\times$ }
\end{center}
   \caption{$4\times$, $16\times$ and $64\times$ upsampling results of \textit{Moai}.
}
\label{m2}
\vspace{-0.3cm}
\end{figure}
\begin{figure}
\begin{center}
{{\hspace{0cm}512 points\hspace{0.43cm}1024 points\hspace{0.43cm}2048 points\hspace{0.43cm}4096 points}}
\includegraphics[width=0.98\linewidth]{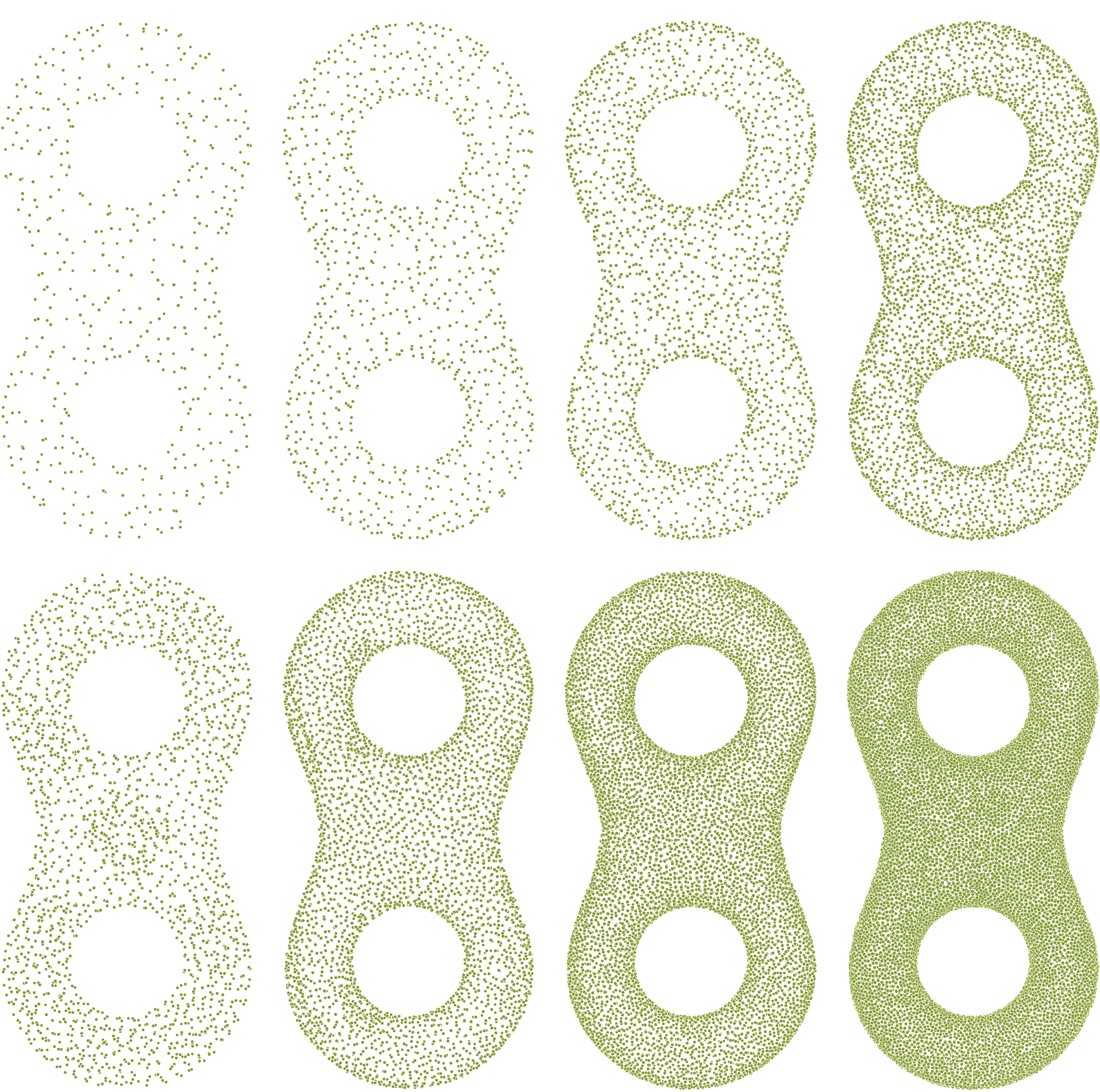}

\end{center}
   \caption{$4\times$ Upsampling results of \textit{Eight} with varying size of input. The first row are the inputs, the second row are the corresponding upsampling results.
}
\label{m3}
\end{figure}

\begin{figure}
\begin{center}
{{\hspace{-0.2cm}Clean\hspace{2.2cm}1\%\hspace{2.15cm}2\%}}
\includegraphics[width=0.98\linewidth]{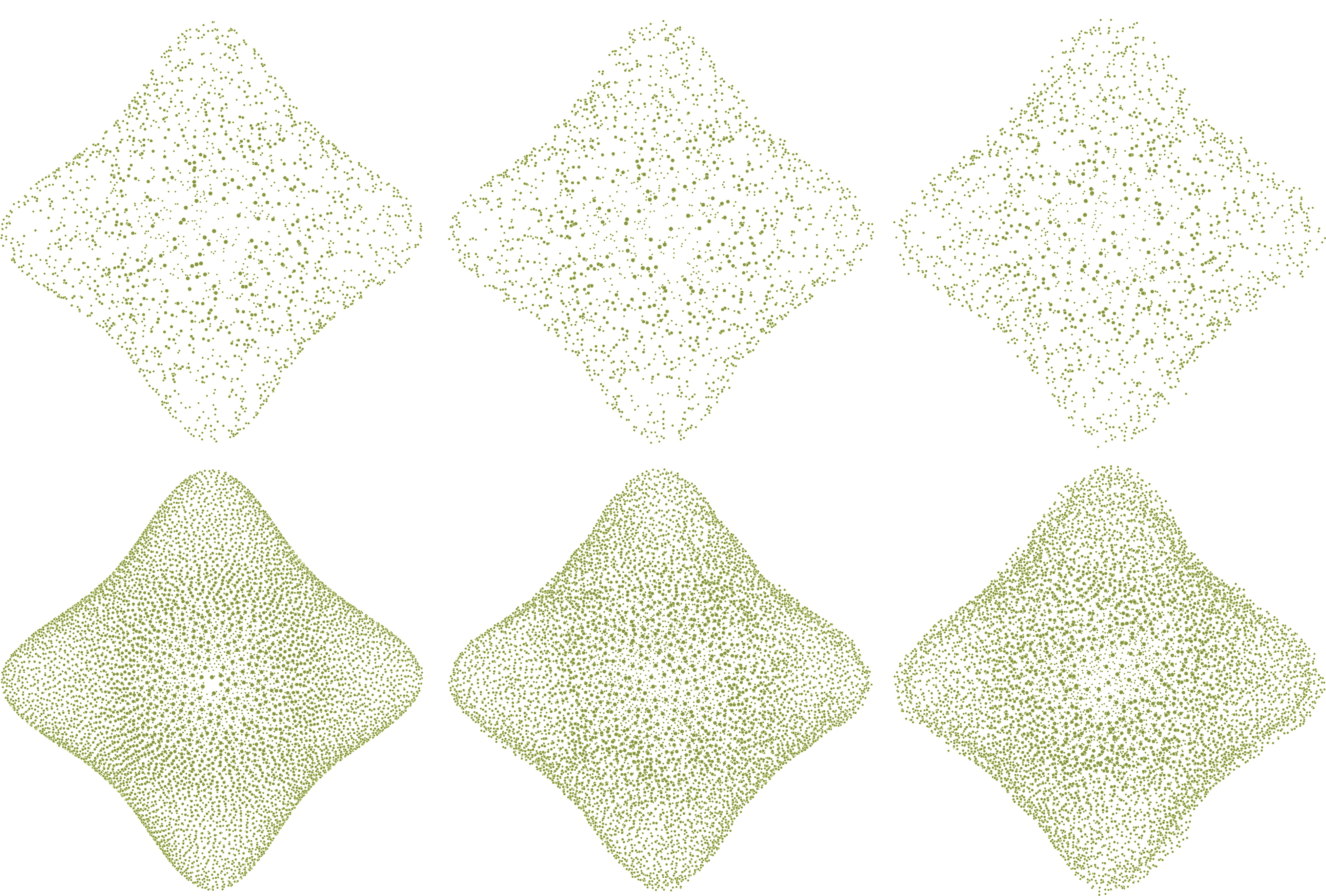}

\end{center}
   \caption{$4\times$ Upsampling results of \textit{Star} with different additive Gaussian noise level.
}
\label{m4}
\vspace{-0.3cm}
\end{figure}

\begin{figure*}
\begin{center}

\includegraphics[width=0.9\linewidth]{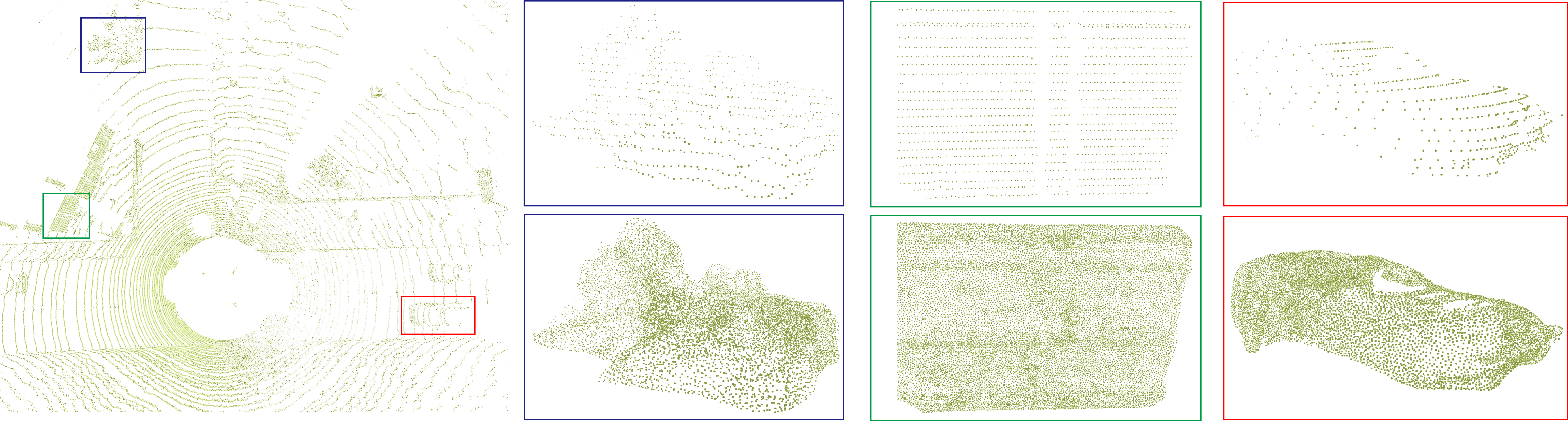}

\end{center}
   \caption{$8\times$ Upsampling result on a real-world sample from KITTI. Please enlarge the PDF for more details.}
\label{m5}
\end{figure*}

\begin{figure}
\begin{center}
\includegraphics[width=0.98\linewidth]{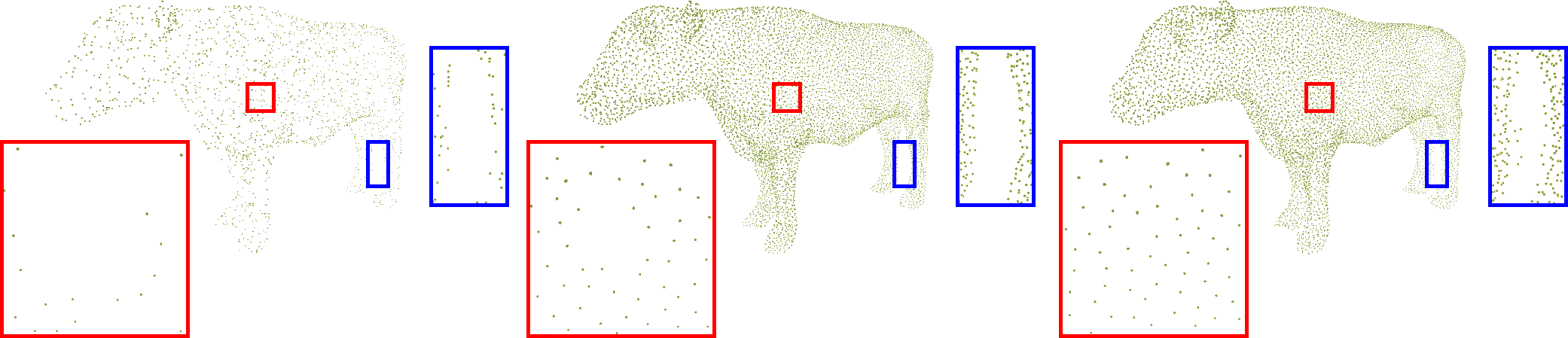}
{{(a)\hspace{2.2cm}(b)\hspace{2.15cm}(c)}}
\end{center}
   \caption{Ablation on the choice of $M$. (a) input point cloud, (b) $M=3$, (c) $M=10$.}
\label{m6}
\end{figure}

\begin{figure}
\begin{center}
\includegraphics[width=0.98\linewidth]{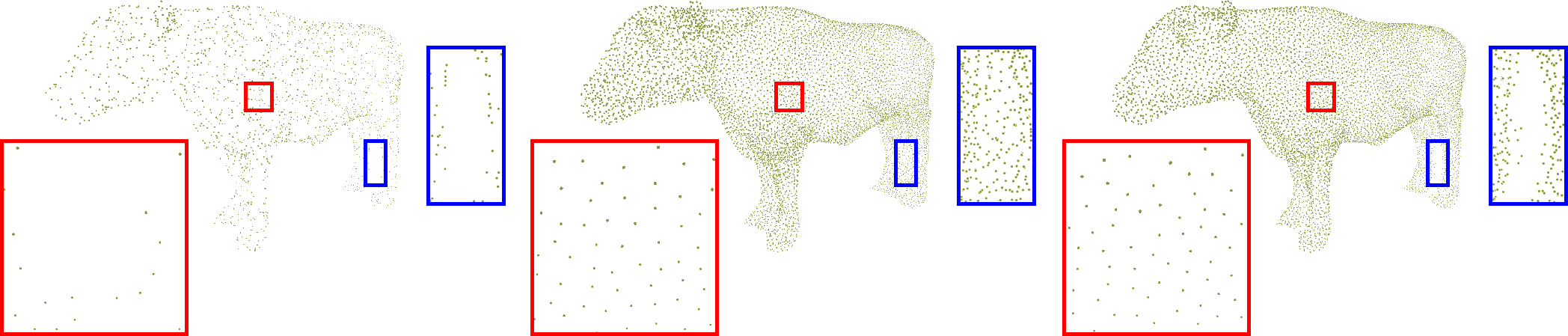}
{{(a)\hspace{2.2cm}(b)\hspace{2.15cm}(c)}}
\end{center}
   \caption{Ablation on the outliers removal. (a) input point cloud, (b) without outliers removal, (c) with outliers removal.}
\label{m7}
\end{figure}

\noindent\textbf{Visual Comparison.} Figure~\ref{m1} illustrates the $4\times$ upsampling results generated by our method and the compared state-of-the-art methods on three models \textit{Chair}, \textit{Camel} and \textit{Fandisk}. The results show that our method  achieves better visual performance than other methods. The produced high-resolution point clouds are dense and uniform, which also have continuous and complete contours. Specifically, results of the highlighted part of \textit{Chair} show that our method succeeds in recovering structure from very few points; results of the highlighted part of \textit{Camel} show that our method can handle complex contour. Furthermore, our method can reconstruct the edge region very well, as demonstrated in the highlighted part of \textit{Fandisk}. The above visual comparisons verify the superiority of our proposed method.

\noindent\textbf{Results on Variable Scales.}  Figure~\ref{m2} shows the upsampling results of \textit{Moai} with different scale factors. 

It can be observed that the contours of all the results are consistent, and the uniformity of points is well preserved.

\noindent\textbf{Robustness against Varying Sizes of Input.}  Figure~\ref{m3} shows the $4\times$ upsampling results of \textit{Eight} with different sizes of input point sets. Our method generates consistent outlines regardless of the number of input points. Figure~\ref{m4} shows the $4\times$ upsampling results of \textit{star} with noise level 0\%, 1\% and 2\%. Our scheme also works well on the noisy input while the uniformity is well preserved. Overall, our method is robust to the input size and noise.

\noindent\textbf{Result on Real-world Sample.} We choose one real-world sample from KITTI~\cite{geiger2012we} to evaluate the generalization capability of our method. In Figure~\ref{m5}, the upsampling results of three different regions are presented. It can be found that, even though the input point cloud is sparse and non-uniform, our scheme can still recover the high-resolution one very well.

\subsection{Ablation Study}

\noindent\textbf{About the choice of $M$.} $M$ works in the distance approximation of a seed point to the underlying surface, which affects the number of outliers and the continuity of projected points. When $M$ increases, both the continuity and the number of outliers increase. To show the effect of different $M$, we provided the 4$\times$ upsampling result of \textit{cow} with $M=3$ and $M=10$. The results are shown in Figure~\ref{m6}. It can be observed that, when $M=3$, no outlier is introduced to the blue box. However, the surface is discontinuous in the red box. When $M=10$, the surface in the red box becomes continuous, however, a few amount of outliers is still introduced to the blue box after outliers removal.

\noindent\textbf{About the necessity of outliers removal.}
To show the effect of outliers removal, we provid the 4$\times$ upsampling result of \textit{cow} with and without outliers removal. The results are shown in Figure~\ref{m7}. 
It can be observed that the outliers removal does not affect the smooth region in the red box, while it can remove most of the outliers in the blue box.

\subsection{Limitations}
The limitation of our method are two-fold. Firstly, even outliers removal is performed, there still exist a certain number of outliers. Secondly, our method cannot precisely control the number of upsampled point set. We have to first generate a dense one with over-sampled points and then adjust the number of vertices to the target number by the farthest point sampling algorithm.

\section{Conclusion}
\label{sec:conclusion}

In this paper, we present a novel and effective point clouds upsampling method via implicit neural representation, which can achieve self-supervised and arbitrary-scale upsampling simultaneously. We formulate point clouds upsampling as the task of seeking nearest projection points on the implicit surface for seed points, which can be done by two implicit neural functions trained without the ground truth dense point clouds. Extensive experimental results demonstrate that our method can produce high-quality dense point clouds that are uniform and complete, and achieves competitive objective performance and even better visual performance compared with state-of-the-art supervised methods.

\section{Acknowledgements}
{This work was supported by National Key Research and Development Project under Grant 2019YFE0109600,  National Natural Science Foundation of China under Grants 61922027, 6207115 and 61932022.}


{\small
\bibliographystyle{ieee_fullname}
\bibliography{egbib}
}

\end{document}